\documentclass[numbers]{article}

\usepackage[preprint, nonatbib]{neurips_2026}
\usepackage[utf8]{inputenc} 
\usepackage[T1]{fontenc}    
\usepackage{hyperref}       
\usepackage{url}            
\usepackage{booktabs}       
\usepackage{amsfonts}       
\usepackage{nicefrac}       
\usepackage{microtype}      
\usepackage{xcolor}         
\usepackage{amsmath}
\usepackage{amssymb}
\usepackage{graphicx}
\usepackage{tabularx}
\usepackage{comment}
\usepackage{cite}

\title{Organization of computation in reservoir computing}

\author{%
  Mohab Abdalla \\
  CentraleSupélec, Université de Lorraine, \\
  LMOPS EA-4423 Laboratory, \\
  57070 Metz, France \\
  \And
  Damien Rontani \\
  CentraleSupélec, Université de Lorraine, \\
  LMOPS EA-4423 Laboratory \\
  57070 Metz, France \\
}

\begin{document}
\maketitle
\begin{abstract}
    Reservoir computing exploits nonlinear dynamical systems to encode temporal inputs into high-dimensional state space representations.
    Although reservoir performance is often characterized through memory, nonlinearity, and their tradeoff, such aggregate measures do not reveal how task-relevant information is organized within the state space.
    Here, we introduce an eigenspectral decomposition framework linking the degree-wise information processing capacity to the corresponding state space modes.
    As a result, we are able to quantify the degree-wise representation energy, and show that in some cases, substantial amounts of information processing capacity may reside in low-energy modes that are vulnerable to experimental noise.
    These results suggest that useful reservoir computation depends not only on dimensionality expansion, but also on the geometric organization of task-relevant information, with direct implications on physical reservoir computers.
       
\end{abstract}


\section{Introduction}
Reservoir computing (RC) is a framework for computing with dynamical systems \cite{jaeger2001, maass2002, nakajima2021reservoir, Cucchi2022} that leverages a system's nonlinear recurrent dynamics, even if they are not completely known, to do useful computation through a projection of the input signal onto a higher dimensional state space.
The RC learns the latent representation of that state space with a decoder, called the readout. 
RC can be thought of as a special case of recurrent neural networks where the input weights and hidden layer weights are fixed and not trained.
Training only the readout weights makes it a simple regression problem, and is made effective (performance-wise) thanks to dimensionality expansion, which improves the probability of finding suitable linear classifiers \cite{cover65}.
This is conceptually similar to feature-space expansion underlying the kernel trick in support vector machines \cite{kernel2025}.
The success of RC is well-documented in the literature, whether in software implementations or physical realizations \cite{Yan2024, Nakajima_2020}.
Physical RC has already been demonstrated across a wide range of platforms, for example in electronics \cite{Liang2024}, photonics \cite{brunner2019photonic, Abdalla2026}, among others \cite{Tanaka2019}. 
Beyond benchmark tasks, several studies have proposed task-independent metrics to quantify overall computing performance in dynamical systems, among them the information processing capacity (IPC) \cite{Dambre2012}.
The IPC quantifies the amount of decodable information present at the linear readout of a stochastically driven causal dynamical system.
This is done by training the readout to reconstruct a set of orthogonal targets defined over the Hilbert space of past inputs.
The IPC measured at the readout is upper bounded by the number of nodes, and more generally by the number of readout observables.
Many studies have used the IPC to probe the overall computational capacity of dynamical systems across different physical platforms \cite{mirko2020, Akashi2020, vettel2022, Ishida2023, Mart2023}.
In addition, multiple works extend the IPC framework to address specific problems or for broader applicability in general \cite{Grigoryeva2015, Kubota2021, SchultetoBrinke2023, Hulser2023, Saito2026}.
However, a missing link is understanding not only how much computation is done by a dynamical system, but also \textit{where} that computation is taking place.
By \textit{where}, we mean the location of task-relevant information within the reservoir state space, i.e., the modes or subspaces that carry features relevant to a given task.
To the best of our knowledge, the way memory and nonlinear transformations are organized within a reservoir's representation of the input history remains largely unexplored.
In this study, we introduce a mode decomposition framework for resolving reservoir state space directions occupied by linear and nonlinear representations, revealing how computation is geometrically organized across the node-space modes and how this organization changes with reservoir parameters.
We perform this analysis with the IPC through projecting the state matrix onto an orthonormal basis and obtaining the per-mode contributions to IPC performance.
Furthermore, we show the degree-wise representation energy, a relevant metric when considering the effective nonlinear capacity of physical reservoir computers.
This approach is applicable beyond task-independent metrics, such as benchmark tasks and real-world applications.
We demonstrate our findings using the standard echo-state network (ESN) model \cite{jaeger2001}.

\section{Results}
\subsection{Decomposition of the state matrix}
\label{gen_inst}

Consider the standard ESN model with \(N\) recurrent nodes. The ESN state vector $\mathbf{x}(t) \in \mathbb{R}^N$ at discrete timestep $t$ is given by 
\begin{equation}
\label{eq1}
\mathbf{x}(t)
=
\tanh\!\left(
\rho\mathbf{W}_{\mathrm{res}}\mathbf{x}(t-1)
+
\iota\mathbf{w}_{\mathrm{in}}u(t-1)
+
\beta \mathbf{b}
\right),
\end{equation}
where $\mathbf{W}_{\mathrm{res}}\in \mathbb{R}^{N\times N}$ is the normalized reservoir connectivity matrix, and $\mathbf{w}_{\mathrm{in}} \in \mathbb{R}^{N}$ is the input-to-reservoir connectivity vector.
Both $\mathbf{W}_\mathrm{res}$ and $\mathbf{w}_\mathrm{in}$ have their entries randomly sampled from \(\mathcal{U}(-1,1)\).
The bias vector is denoted by $\mathbf{b}$ with entries \(b_i \overset{\mathrm{i.i.d.}}{\sim} \mathcal{U}(-1,1)\). The input \(u(t-1)\) is single channel, and
\(\rho\), \(\iota\), and \(\beta\) are the spectral radius, input scaling factor, and bias scaling factor, respectively.

We inject a single-channel white noise input sequence of length $T$
\(
\{u(t)\}_{t=1}^{T},
\)
with samples drawn randomly as \(
u(t) \overset{\mathrm{i.i.d.}}{\sim} \mathcal{U}(-1,1)
\).
The reservoir state matrix is the concatenation of all state vectors \(\mathbf{x}(1)\) to \(\mathbf{x}(T)\) resulting in
\(
\mathbf{X} \in \mathbb{R}^{N \times T}.
\)
The singular value decomposition (SVD) of \( \mathbf{X}\) is given by
\(\mathbf{X}
=
\mathbf{Q}\boldsymbol{\Sigma}\mathbf{V}^{\top}.
\)
Here, \(\mathbf{Q} \in \mathbb{R}^{N\times N}\) contains the left singular vectors, forming an orthonormal basis in the node space.
The vector of singular values \(\boldsymbol{\sigma}=[\sigma_1,\sigma_2,\ldots,\sigma_r]^{\top}\), whose elements lie on the diagonal of \(\boldsymbol{\Sigma}\), quantifies the variance, or "strength", of each mode.
Throughout this work, we keep the full vector of singular values \(\boldsymbol{\sigma}\), i.e. \(r=N\).
\(\mathbf{V}^{\top}\) contains the corresponding temporal modes describing how each node-space mode evolves over time.
The energy fraction carried by mode \(i\) is
\begin{equation}   
\eta_i
=
\frac{\sigma_i^2}
{\sum_{j=1}^{r} \sigma_j^2}.
\end{equation}
\begin{figure}[b]
    \centering
    \includegraphics[width=\linewidth]{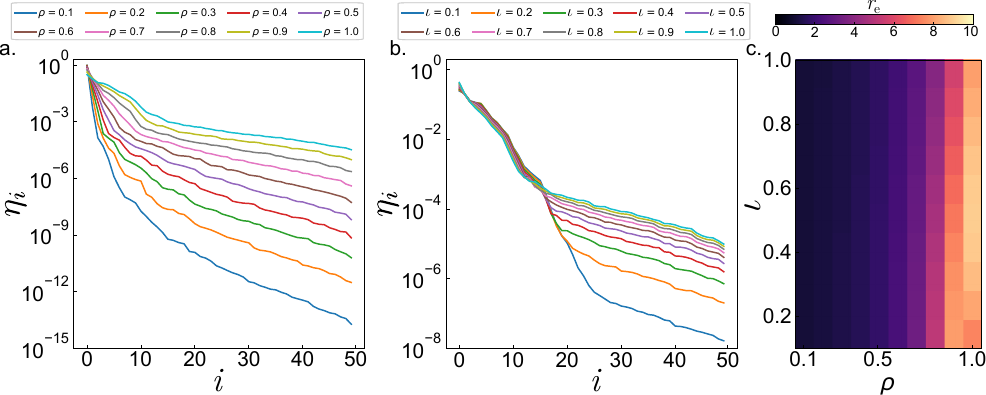}
    \caption{Energy fractions for different values of (a) spectral radius with \(\iota=1\) and (b) input scaling factors with \(\rho=0.5\). In (c), the effective rank is shown for varying \(\rho\) and \(\iota\). }
    \label{fig1}
\end{figure}

We plot these energy fractions for a sweep of the spectral radius and the input gain, and \(\beta=0\), as shown in Fig. \ref{fig1}.
In panel a, we observe that increasing the spectral radius has an effect of increasing the dimensionality of the reservoir activity.
For example, with \(\rho=0.1\) (\(\rho=0.9\)), \(\eta_{49}\approx10^{-14}\) (\(\eta_{49}\approx10^{-5}\)).
This is expected as we move from more contractive dynamics in smaller \(\rho\) to richer and more prolonged dynamics in larger \(\rho\).
In this context, the `flatter' distribution of the energy fractions for larger \(\rho\) means more dimensions are being utilized effectively in the state space.
In Fig. \ref{fig1}b, we observe that increasing the input gain seems to affect in particular the `tail' modes.
We can see from both panels that the majority of reservoir activity is dominated by a few modes.
This is consistent with the observation that reservoir activity is effectively low-dimensional compared to \(N\) \cite{jaeger2014}.
The effective dimensionality of the state space can be measured by the effective rank \(r_\mathrm{e}\) \cite{Olivier2007}
\begin{equation}
    r_\mathrm{e} = \exp\left( -\sum_{i=1}^r \eta_i \ln(\eta_i) \right).
\end{equation}
As shown in Fig. \ref{fig1}c, it is observed that the maximum \(r_\mathrm{e}\approx 9\) for a reservoir of size \(N=50\). 
Furthermore, it can be observed that for the chosen parameters, \(r_\mathrm{e}\) is more sensitive to \(\rho\) than \(\iota\).
This naturally raises the question of which modes are being occupied by the relevant task features, and is the motivation for applying the mode-task projection framework to the IPC, as discussed in the following section.

\subsection{Mode-task projection}
\label{sec:modetask}
In this section, we propose a method to view the IPC from a state space geometry perspective. 
We briefly revisit the definition of IPC before introducing the mode-task projection framework.
The IPC, introduced in \cite{Dambre2012}, is defined with respect to an infinite space of orthogonal targets \(\mathcal{Z}\) which are constructed from products of Legendre polynomials of delayed versions of the input sequence
\begin{equation}
z_k(t)
=
\prod_m \mathcal{P}_{d_m}\!\left(u(t-m)\right),
\end{equation}
where \(\mathcal{P}_{d_m}\) is a polynomial of degree \(d\) applied on the \(m\) steps delayed input, and \(k\) indexes a particular choice of the degrees across memory delays $\{d_m\}$.
Legendre polynomials are typically used, but other polynomial families can be considered as well for different input distributions \cite{Kubota2021}.
The input sequence's elements are sampled i.i.d. from \( \mathcal{U}(-1,1)\) to construct orthogonal target vectors \(\mathbf{z} \in \mathcal{Z}\). 
The degree \(d\) of a target refers to the degree of the corresponding polynomial product. 

Consider the subspace $\mathcal{Z}_d$ of targets produced by degree $d$ products of Legendre polynomials. 
As discussed in \cite{Dambre2012}, there exists some subset of $\mathcal{Z}_d$ which contains those finite-dimensional target vectors that can be reconstructed with a capacity beyond some threshold \(\varepsilon\)
\begin{equation}
\mathcal{Z}_{d}^{(\varepsilon)}
    =
\left\{
\mathbf{z}_{k}\in \mathcal{Z}_{d}
\;\middle|\;
c_k>\varepsilon
\right\}
\subseteq
\mathcal{Z}_{d},
\end{equation}

with \(0\leq c_k \leq 1\) being the capacity to reconstruct some target vector \(\mathbf{z}_k\) indexed by \(k\), and in the case of ordinary least squares is given by
\begin{equation}
\label{eq5}
c_k
=
\frac{\left\|\mathbf{VV}^{\top} \mathbf{z}_k\right\|^2}
{\left\|\mathbf{z}_k\right\|^2},
\end{equation}
where \(\|\cdot\|^2\) denotes the squared Euclidean norm. 
This formulation is similar to the one presented in \cite{Kubota2021}.
Summing up for all \(k\) yields the IPC
\begin{equation}
    C = \sum_k c_k.
\end{equation}
Using the node-space basis \(\mathbf{Q}\), we can obtain the projection of the states onto each node-space mode \(i\)
\begin{equation}
\mathbf{\tilde{x}}_i = \mathbf{q}_i^{\top}\mathbf{X},
\end{equation}
where \(\mathbf{q}_i \in \mathbb{R}^{N}\) is the \(i\)-th basis vector, and \(\mathbf{\tilde{x}}_i \in \mathbb{R}^{1\times T}\) denotes the reservoir state projected onto the \(i\)-th node-space mode \(\mathbf{q}_i\).
These projected states can equivalently be obtained from the rows of
\begin{equation}
    \widetilde{\mathbf{X}} = \mathbf{Q}^{\mathsf{T}}\mathbf{X}.    
\end{equation}
From a physical perspective, \(\mathbf{\widetilde{X}}\) are the reservoir dynamics expressed in the orthonormal basis \(\mathbf{Q}\) of the node-space.

Given a set of target sequences, consider the task of reconstructing a target \(\mathbf{z}_k \in \mathbb{R}^T\).
We can measure how much each mode \(i\) contributes to reconstructing \(\mathbf{z}_k\) using the squared cosine similarity between the two vectors (hereafter referred to as the mode-task score) 
\begin{equation}
s_{i,k}
=
\frac{\left(\mathbf{\tilde{x}}_{i}{}^{\top}\mathbf{z}_{k}\right)^2}
{\left\|\mathbf{\tilde{x}}_{i}\right\|^2 \left\|\mathbf{z}_{k}\right\|^2}.   
\end{equation}
Formally, \(s_{i,k}\) is defined in a common inner-product space containing both the projected state components of \(\widetilde{\mathbf{X}}\) and the target vectors in \(\mathcal{Z}\).
It quantifies the alignment between the $i$-th state space mode and the $k$-th target task.
As a result, the mode-task score is bounded as \(0 \leq s_{i,k} \leq 1\) by the Cauchy–-Schwarz inequality \cite{Horn_Johnson_1985}. 
It is also useful to look at a cumulative picture of the mode-task score.
The cumulative mode-task score up to mode \(i\) is given by
\begin{equation}
    S_k^{i} = \sum_{j=0}^{i} s_{j,k},
\end{equation}
with \(j\) being a dummy variable for the summation.
This quantity is useful to indicate the onset and offset of performance for a given task, i.e. `where' a task starts and stops being represented.
The onset mode \(i\) is the first mode where \(S_k^{i}\ge \delta_\mathrm{on}\) and \(S_k^{i-1}<\delta_\mathrm{on}\), where \(\delta_\mathrm{on}\) is the onset threshold.
The offset mode is defined where \(S_k^{i+1}-S_k^{i}<\delta_\mathrm{off}\).  
This can be used to quantify the number of dimensions being utilized to solve a given task, independent of the performance.
Summing for all \(i\) gives the capacity to reconstruct task \(k\)
\begin{equation}
  S_k = \sum_{i=0}^{N-1} s_{i,k}  = c_k,
\end{equation}
we can thus update Equation \ref{eq5} to the following form
\begin{equation}
\label{eq11}
    c_k
    =
    \frac{\left\| \mathbf{V}\mathbf{V}^{\top}\mathbf{z}_k \right\|^2}
    {\left\| \mathbf{z}_k \right\|^2}
    =
    \sum_{i=0}^{N-1}
    \frac{
    \left( \tilde{\mathbf{x}}_i\mathbf{z}_k \right)^2
    }{
    \left\| \tilde{\mathbf{x}}_i \right\|^2
    \left\| \mathbf{z}_k \right\|^2
    }.
\end{equation}
\begin{figure}[b]
    \centering
    \includegraphics[width=\linewidth]{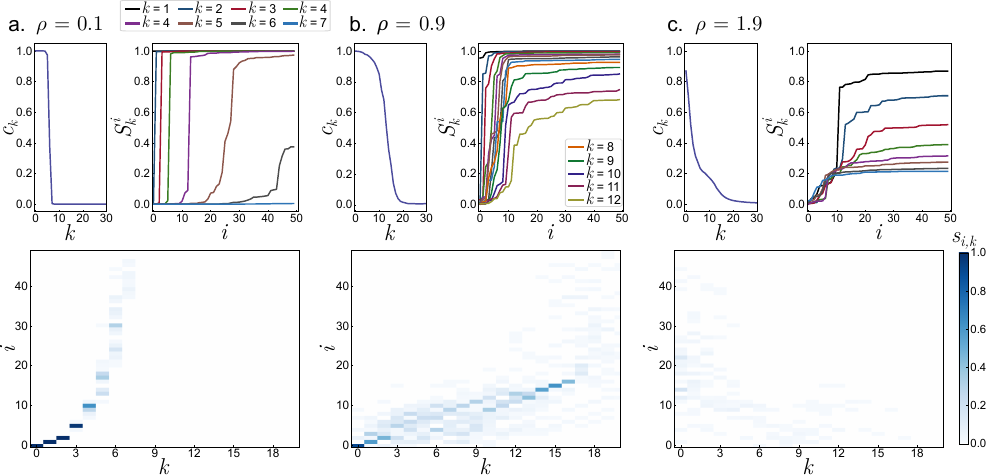}
    \caption{The mode task-projection is shown for \(d=1\) for three different cases of spectral radius. 
    For each panel: the capacity to reconstruct the task \(k\) (top left) is decomposed into the mode basis (bottom).
    The cumulative mode-task scores (top right) shows the onset and level-off of the task performance in terms of the modes being recruited.}
    \label{fig2}
\end{figure}

The left-hand side corresponds to the standard IPC picture, formulated in a manner similar to \cite{Kubota2021}, where capacity is interpreted as the normalized projection energy of the target onto the temporal subspace spanned by the reservoir states. The right-hand side rewrites the same quantity as a sum of mode-resolved contributions associated with the node-space modes.
In finite-length simulations, small discrepancies may arise between the two quantities, primarily due to empirical non-orthogonality of the target vectors.
We have observed that these small discrepancies decrease for longer input sequence lengths, as expected.
Similarly, we can define the usage of mode \(i\) \(\forall~\mathbf{z}_k \in \mathcal{Z}^{(\varepsilon)}\) as
\begin{equation}
    S_i = \sum_{k} s_{i,k}.
\end{equation}
For finite input sequences, the upper bound of the mode usage is only approximate, so we write \(0\leq S_i\lesssim1\). 
The upper bound is exactly \(S_i \leq 1 \; \text{as} \; T \to \infty\) (see Appendix).

Consider the tasks of the first degree of IPC ($d=1$), i.e. the linear memory capacity \cite{Jaeger_2001}. 
The targets are the memory functions of the input sequence, and the task is to reconstruct the input \(m\) memory steps into the past, i.e. 
\(
{z}_k(t) = u(t-m)
\).
In this case, the targets are indexed by $k$ starting from 0, while \(m\) starts from 1, so that \(k=m-1\).
In Fig. \ref{fig2}, we can observe how each task is decomposed into its mode projections. 
In other words, we reveal which reservoir modes contain a task’s relevant features.
For each panel in Fig. \ref{fig2}, the top left subpanel is the standard memory function \(c_k\), the top right is the cumulative mode-task score \(S_k^i\), and the bottom subpanel is the heatmap for \(S_{i,k}\). 
We show results for three different cases of spectral radius, corresponding to three dynamical regimes: highly stable (\(\rho=0.1\)), near-critical (\(\rho=0.9\)), and unstable (\(\rho=1.9\)).
In Fig. \ref{fig2}a we can observe that for \(\rho=0.1\) and for all \(k\), \(S_k^i\) seem to be dispersed, with wide spacings between the onsets of the different tasks \(k\).
This suggests a small amount of mode sharing for the different tasks' features.
The associated \(s_{i,k}\) are strongly localized for the first few tasks.
For example, tasks \(k=0\) to \(k=3\) are employing almost exclusively features embedded within a single dynamical mode. 
Conversely, for \(\rho=0.9\) (Fig. \ref{fig2}b), it seems most of the modes are contributing to multiple tasks simultaneously as shown in the corresponding \(s_{i,k}\) heatmap. 
In addition, almost all the tasks recruit features from multiple modes.
In panel \ref{fig2}c, with \(\rho=1.9\) corresponding to the unstable regime, the \(S_k^i\) onset is shifted to higher-index modes.
This can be explained as a small but useful portion of the signal buried under highly unstable dynamics, which occupy the dominant modes.
In general, given a reservoir with a certain configuration under some input excitation, each task can be thought of as having a specific “fingerprint” on the basis \(\mathbf{Q}\). 

\subsection{Organization of computation}
In this section, we use the mode usage quantity defined in the previous section to determine \textit{where} the IPC tasks, grouped by degree, live in the reservoir's modes.
That is to say, if a reservoir's dynamical space is, for example, an $N$-dimensional ellipsoid, we wish to examine which parts of that ellipsoid is being used in the given task. 
We use the mode usage per degree to quantify the sum contribution of all tasks of degree d to the mode usage. This can be written as
\begin{equation}
    S_{i,d} = \sum_{k|\mathbf{z}_k \in \mathcal{Z}_d^{(\varepsilon)}}
    s_{i,k}.
\end{equation}

In the calculation of IPC, specifically in the SVD step, we choose a small threshold of absolute singular values of \(10^{-10}\) for negligible rank truncation. 
In our results, this chosen value retained all \(N\) modes in the reservoir.
This is not to be confused with the capacity threshold \(\varepsilon\) described in section \ref{sec:modetask}.

\begin{figure}[b]
    \centering
    \includegraphics[width=\linewidth]{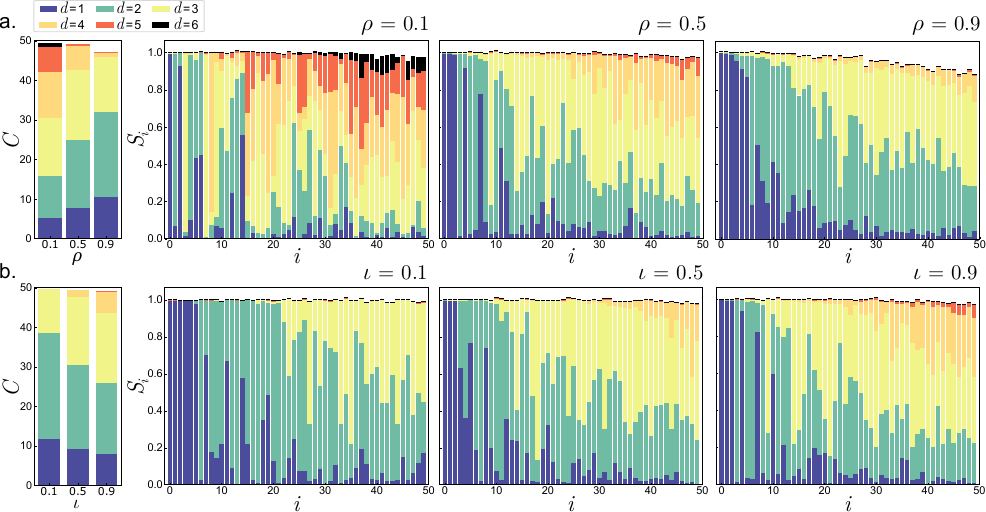}
    \caption{The IPC, shown in the leftmost panels, is decomposed into the mode usage for different values of a. spectral radius with \(\iota=1.0\) and b. input gain with \(\rho=0.5\).}
    \label{fig3}
\end{figure}

As shown in Fig. \ref{fig3}a, the leftmost panel shows the IPC for three reservoirs with \(\rho=\{0.1, 0.5, 0.9 \}\), \(\iota=1.0, \beta=0.5\).
For each \(\rho\), we plot the mode usage across all the modes.
We stack the corresponding \(S_{i,d}\), which yields \(S_{i}\). 
With \(\rho=0.1\), the degree contributions seem dispersed across the modes.
By increasing \(\rho\), a particular organization begins to appear.
Low-index (higher-variance) modes seem to be occupied mainly by linear memory features, while higher-index modes seem to encode nonlinear transformations.
Furthermore, the onset of the different degrees happens sequentially.
The first degree (blue) appears before the second (green), and the second before the third (yellow), and so on.
This suggests that higher order nonlinear interactions tend to be embedded in modes with smaller variance in ESNs.
From a physical point of view, this is not surprising, especially for fading memory systems where the influence of inputs deeper in the past is much lower than recent inputs. 
Typically, the longer surviving portions of the input history are the ones that experience more nonlinear interactions.

Similarly, we show the results for three cases of input scaling factor \(\iota=\{0.1, 0.5, 0.9\}\). 
Increasing \(\iota\) means that a larger range of the activation function is utilized, thereby increasing the amount of nonlinear transformation.
Indeed, the stronger input injection results in higher degree capacities appearing, as shown in the IPC plot of Fig. \ref{fig3}b. 
However, the mode usage also shows that higher degree capacities are pushed to occupy more of the higher-variance (lower-index) modes, at the expense of lower degree capacities.
This suggests a potentially more accessible representation for higher degree capacities.
For all the cases studied in this section, it seems that the first mode or few modes remain dominated by linear memory \((d=1)\).
The results presented here raise the question of how much of the signal energy is being used to represent degree $d$ tasks' features, which is addressed in the next section.
\begin{table}[b]
    \centering
    \renewcommand{\arraystretch}{1.25}
    \begin{tabular}{|c|c|c|c|c|c|c|}
    \hline
        & \multicolumn{3}{c|}{Biased \(\tanh\)} 
        & \multicolumn{3}{c|}{Unbiased \(\tanh\)} \\
    \hline
        Quantity 
        & \(\mu \pm \sigma\) & min & max
        & \(\mu \pm \sigma\) & min & max \\
    \hline
        \(\Phi_1\) 
        & \(0.997 \pm 3.50{\times}10^{-3}\) & 0.947 & 1.005
        & $0.999 \pm 2.3\times10^{-3}$ & 0.864 & 1.01 \\
        
        \(\Phi_2\) 
        & \(2.29{\times}10^{-3} \pm 2.75{\times}10^{-3}\) & $0.0002$ & 0.033
        & $\leq 10^{-4}$ & $\leq 10^{-4}$ & $0.0005$ \\
        
        \(\Phi_3\) 
        & \(1.45{\times}10^{-3} \pm 1.25{\times}10^{-3}\) & $0.0002$ & 0.026
        & $0.002 \pm 0.0019$  & $0.0004$ & 0.041  \\
        
        \(\Phi_4\) 
        & \(7.32{\times}10^{-4} \pm 2.93{\times}10^{-4}\) & $0.0001$ & 0.003
        & $2.7 \times 10^{-4}\pm 10^{-4}$ & $\leq 10^{-4}$ & $0.0009$ \\
        
        \(\Phi_5\) 
        & \(7.46{\times}10^{-4} \pm 2.50{\times}10^{-4}\) & $0.0002$ & 0.002
        & $1.14 \times 10^{-4} \pm 4\times 10^{-4}$ & $0.0002$ & 0.003 \\
        
        \hline
        \(\bar{C}_1\) & \(0.178 \pm 0.047\)  & 0.097 & 0.501 
        & $0.210 \pm 0.060$ & 0.127 & 0.764 \\
        \(\bar{C}_2\)  & \(0.36 \pm 0.085 \) & 0.176 & 0.659
        & 0.0 & 0.0 & 0.0\\
        \(\bar{C}_3\)  & \(0.35 \pm 0.068 \) & 0.009 & 0.578 
        & $0.677 \pm 0.084$ & 0.236 & 0.784 \\
        \(\bar{C}_4\)  & \(0.084 \pm 0.073 \) & 0.0 & 0.251 
        & 0.0 & 0.0 & 0.0 \\
        \(\bar{C}_5\)  & \(0.023 \pm 0.036 \) & 0.0 & 0.181 
        & $0.112 \pm 0.12$ & 0.0 & 0.434\\
    \hline
    \end{tabular}
    \caption{Mean, standard deviation, minimum, and maximum of total normalized energy used by degree \(d\) tasks, and normalized capacities per degree, for biased and unbiased $\tanh$ activation functions.}
    \label{tab1}
\end{table}
\begin{table}[t]
    \centering
    \renewcommand{\arraystretch}{1.25}
    \begin{tabular}{|c|c|c|c|c|c|c|}
    \hline
        & \multicolumn{3}{c|}{Biased \(\sin^2\) at \(\pi/4\)} 
        & \multicolumn{3}{c|}{Unbiased \(\sin^2\)} \\
    \hline
        Quantity 
        & \(\mu \pm \sigma\) & min & max
        & \(\mu \pm \sigma\) & min & max \\
    \hline
        \(\Phi_1\) 
        & \(0.99 \pm 5.74{\times}10^{-3}\) & 0.577 & 1.004
        & $0.789 \pm 0.21$ & 0.023 & 0.995 \\
        
        \(\Phi_2\) 
        & $<10^{-4}$ & $<10^{-4}$ & 0.0005
        & $0.21 \pm 0.21$ & $0.005$ & $0.977$ \\
        
        \(\Phi_3\) 
        & \(2.48{\times}10^{-3} \pm 2.95{\times}10^{-3}\) & $0.0004$ & 0.044
        & $0.0019 \pm 0.0017$  & $0.0002$ & 0.021  \\
        
        \(\Phi_4\) 
        & \(2.68{\times}10^{-4} \pm 10^{-4}\) & $<10^{-4}$ & 0.0009
        & $0.001\pm 3.48\times10^{-4}$ & $0.0002$ & $0.0043$ \\
        
        \(\Phi_5\) 
        & \(1.18{\times}10^{-4} \pm 4.38{\times}10^{-4}\) & $0.0003$ & 0.003
        & $0.001 \pm 2\times 10^{-4}$ & $0.0002$ & 0.0024 \\
        
        \hline
        \(\bar{C}_1\) & \(0.20 \pm 0.052\)  & 0.12 & 0.87 
        & $0.094 \pm 0.021$ & 0.054 & 0.192 \\
        \(\bar{C}_2\)  & \(0.0 \) & 0.0 & 0.0
        & \(0.242 \pm 0.066 \) & 0.121 & 0.522 \\
        \(\bar{C}_3\)  & \(0.692 \pm 0.083 \) & 0.13 & 0.79 
        & $0.340 \pm 0.054$ & 0.151 & 0.444 \\
        \(\bar{C}_4\)  & \(0.0 \) & 0.0 & 0.0 
        & \(0.246 \pm 0.054\) & 0.0008 & 0.324 \\
        \(\bar{C}_5\)  & \(0.106 \pm 0.116 \) & 0.0 & 0.44 
        & $0.069 \pm 0.067$ & 0.0 & 0.271\\
    \hline
    \end{tabular}
    \caption{Mean, standard deviation, minimum, and maximum of total normalized energy used by degree \(d\) tasks, and normalized capacities per degree, for biased and unbiased $\sin^2$ activation functions.}
    \label{tab2}
\end{table}
\subsection{Memory–nonlinearity energy allocation}
Consider mode \(i\) with associated energy fraction \(\eta_i\) and usage \(S_i\).
We can write the energy-normalized contribution per degree as 
\begin{equation}
    \phi_{i,d}=\eta_i S_{i,d}.
\end{equation}
In the case of \(r=N\), summing over all the modes yields how much of the normalized total energy in the signal is being used to represent degree \(d\) tasks
\begin{equation}    
\Phi_d=\sum_{i=0}^{N-1}\phi_{i,d}
=\sum_{i=0}^{N-1}\eta_i S_{i,d}
\end{equation}
We refer to this quantity as the normalized degree-wise representation energy, or representation energy for short. For the results reported in this section, we considered \(10000\) random initializations of ESNs with \(N=50\), sampled from the parameter space \(\Theta\) specified in Equation \ref{eq_params}, \(\kappa\) being the sparsity of \(\mathbf{W}_\mathrm{res}\). 
\begin{equation}
\label{eq_params}
\Theta =
\left\{
\begin{aligned}
&\rho \sim \mathcal{U}(0.1, 1.0), \\
&\iota \sim \mathcal{U}(0.1, 1.0), \\
&\beta \sim \mathcal{U}(0.1, 0.5), \\
&\kappa \sim \mathcal{U}(0,0.9), \\
&\mathbf{W}_{\mathrm{res}} \sim \mathcal{U}(0,1)^{N \times N}, \\
&\mathbf{w}_{\mathrm{in}} \sim \mathcal{U}(-0.25,0.75)^{N}, \\
&\mathbf{b} \sim \mathcal{U}(-1, 1)^{N}, \\
&\mathbf{u} \sim \mathcal{U}(-1, 1)^{T}
\end{aligned}
\right\}.
\end{equation}
For each initialization, \(\Phi_d\) is calculated for degrees 1 through 5.
Table \ref{tab1} shows the mean and standard deviation of \(\Phi_d\).
In addition, we consider the normalized capacities per degree as shown in Equation \ref{eqnorm} for a direct comparison between the degree-wise IPC and the corresponding representation energy.
\begin{equation}
\label{eqnorm}
\begin{aligned}
\bar{C}_d = \frac{C_d}{\sum_d C_d},
\quad\bar{C}=
\sum_d \bar{C}_d = 1.
\end{aligned}
\end{equation}

As shown in Table \ref{tab1}, with \(\tanh\) activation, an interesting observation is that tasks with \(d=1\) (associated with linear memory) seem to dominate the energy usage with on average \(\Phi_1 = 0.997\), while expressing an average \(\bar{C}_1 = 0.178\) of the IPC.
This result helps to distinguish between absolute computational capacity, as measured by IPC, and the representation of degree \(d\) tasks in the reservoir's response.
Large capacities may exist in a small representational energy, and vice versa.
For example, as in the case of biased \(\tanh\), \(\bar{C}_2=0.36\) but \(\Phi_2=2.29\times 10^{-3}\). associated with tasks of some nonlinear degree \(d\), but it may be represented in weaker modes.
On the other hand, a smaller \(\bar{C}_1=0.178\) is represented in \(\Phi_1=0.997\).
We also show the unbiased case, where 
\(\mathbf{b}=\mathbf{0}\).
As expected, the corresponding \(\Phi_d\) \((\bar{C_d})\) for even degrees are small (0.0), due to the odd symmetry of the unbiased $\tanh$.

In the case of \(\sin^2\) activation, we consider the fixed biased case where
\(
\mathbf{b} = \frac{\pi}{4}\mathbf{1}_N, 
\)
where \(\mathbf{1}\) is the all-ones vector, and the unbiased case where \(\mathbf{b}=\mathbf{0}\).
The results are reported in Table \ref{tab2}.
The \(\pi/4\) biased \(\sin^2\) resembles the unbiased \(\tanh\) due to the odd symmetry.
This is ensured by restricting the input scaling factor in Equation \ref{eq_params} to \(\iota\in[0,0.5]\) for this case.
Interestingly, not only are the normalized capacities close in both cases, but also \(\Phi_d\).
We also observe strong differences between the unbiased and biased \(\sin^2\) for both quantities.

In the case of biased tanh, we allow \(\mathbf{b}\) to take any value as indicated by Equation \ref{eq_params}. 
The operation regimes of the biased \(\tanh\) are different to the unbiased \(\sin^{2}\). 
Comparing both is still meaningful to illustrate the notion of nonlinear degree accessibility.
Consider the case of \(\Phi_2\) for both activations. 
We can see that for the \(\tanh\) case, \(\bar{c}_2=0.36\) on average, represented by an average \(\Phi_2=2.29\times10^{-3}\). 
Here, we can see that there is indeed a large degree 2 computational capacity, albeit weakly represented in the reservoir state space.
On the other hand, in the unbiased \(\sin^2\) case, \(\bar{c}_2=0.24\) on average, which is less than the \(\tanh\) case. 
However, this amount is represented by \(\Phi_2=0.21\) on average.
This has direct consequences on physical reservoirs, where the unbiased \(\sin^2\) case seems to offer accessibility to second degree nonlinearity in the presence of setup noise.
In the following section, we proceed to discuss further the implications on physical reservoirs, especially with respect to the impact of additive noise.

\subsection{Implications on physical RC}
While it may be considered that an RC with small \(\Phi_d\) but large \(\bar{C}_d\) is computationally efficient, we rather consider the pertinent case of physical RC and its general limitations, specifically the presence of noise.
For insufficient \(\Phi_d\), the direct implication is that accessing higher order nonlinear capacities may be limited, where setup noise can limit the accessibility to higher order transformations.
The degrading impact of noise on the memory capacity and IPC of reservoirs is well-known and has been reported in several works whether numerically or experimentally \cite{Jaeger_2001, vettel2022}. 
The framework presented here provides a degree-resolved view which can be used to further elucidate that impact.

As an example, consider the case of additive white noise in a reservoir computer. We can express the contributions of that noise in terms of allocated energy \(\Phi_\mathrm{noise}\) of the noisy state matrix, representing the effective noise floor.
The SNR per degree can thus be written as
\begin{equation}
    \gamma_d
    =
    \frac{\Phi_d}{\Phi_{\mathrm{noise}}},
\end{equation}
from which we construct a noise-aware degree-wise capacity
\begin{equation}
\bar{C}_d^\gamma
=
\bar{C}_d
\frac{\Phi_d}{\Phi_d+\Phi_{\mathrm{noise}}}
=
\bar{C}_d
\frac{\gamma_d}{1+\gamma_d}.
\end{equation}
This quantity is bounded as \(0\leq \bar{C}^\gamma_d\leq 1\).
Thus, when \(\Phi_d\gg \Phi_\mathrm{noise}\), \(\bar{C}^\gamma_d\approx\bar{C}_d\), i.e. the capacity of degree \(d\) is almost fully recoverable, and when \(\Phi_d \ll \Phi_\mathrm{noise}\), a small or negligible portion of degree \(d\) capacity survives.
When the corresponding representation energy is at the noise floor, \(\Phi_d=\Phi_\mathrm{noise}\), and \(\bar{C}^\gamma_d=\bar{C}_d/2\).
This should not be interpreted as meaning that half of the capacity remains physically recoverable.
Rather, this represents the amount of imposed penalty that suppresses capacity contributions whose associated representation energy is comparable to the noise floor.
This can provide insights into task performance of a physical RC system with respect to varying SNR, as previously investigated in some earlier studies \cite{Vinckier15, Duport2016}. 
Assuming isotropically distributed additive noise, and assuming that the clean reservoir state matrix has an effective rank sufficiently smaller than the full state dimension \(N\), there exists at least one singular direction whose normalized energy is strongly dominated by noise. Formally,
\begin{equation}
r_{\mathrm{e}}(X) < N
\quad \Rightarrow \quad
\exists i \in \{0,\ldots,N-1\}
\ \text{such that}\
\eta_{i,\mathrm{signal}} \ll \eta_{i,\mathrm{noise}}.
\end{equation}
In practice, one can use the weakest mode effectively as a noise-only direction,
\begin{equation}
    \eta_{\mathrm{noise}} \approx \eta_{N-1}.
\end{equation}

\begin{figure}[t]
    \centering
    \includegraphics[width=\linewidth]{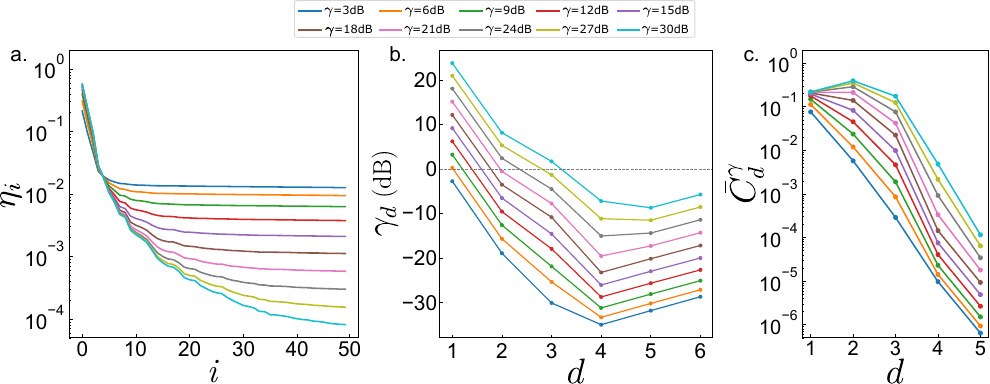}
    \caption{The impact of varying the input SNR \(\gamma\) on the different degrees of IPC. (a) Energy fractions of the noisy state matrices. The setup is similar to Fig. \ref{fig1}a, however, here $\beta$=0.5. (b) \(\gamma_d\) expressed in dB, (c) the noise-aware degree-wise capacity. Throughout the three panels, the noise strength is directly expressed in terms of the input SNR \(\gamma\).}
    \label{fig4}
\end{figure}

Under the assumption of isotropy, the total normalized noise contribution can be roughly estimated as
\begin{equation}
\label{eq:noisefloor}
\Phi_{\mathrm{noise}}
\approx N \eta_{\mathrm{noise}}.
\end{equation}
In Fig.~\ref{fig4}, we show the impact of readout noise on the different degrees of IPC on a tanh ESN. 
In panel a, we can observe the noise floor in the approximate flattening of the spectrum after some index \(i\). 
We choose \(\eta_{49}\) to calculate \(\Phi_\mathrm{noise}\) for each case of \(\gamma\).
We plot \(\gamma_d\) for each degree.
By using 0 dB (dotted gray line) as the reference, which indicates that the signal is as much as the noise, we can track which capacities survive depending on the readout noise strength.
For example, when the noise strength is low (\(\gamma=30\) dB), we can see that for \(d=1,2,3\), \(\gamma_d\) is above the noise level.
However, in the case of \(\gamma=3\) dB, much of \(\gamma_d\) is buried deeply under the noise, especially for \(d\geq2\).
While this does not necessarily mean the complete destruction of computation ability, it severely impacts the amount of recoverable features from the state space.
Therefore, the reservoir may generate higher-degree representations in the noiseless case, but most of them are not robust enough to remain distinguishable under realistic noise considerations.
This is elucidated by the noise-aware degree-wise capacity, shown in Fig. \ref{fig4}c.
We can see that the normalized capacities quickly drop as a function of increasing \(d\).
This means that it is consistently harder for higher degrees to survive the impact of noise in a tanh ESN.
We also remark that the sensitivity of degree \(d\) capacities is linked to the representation energy \(\Phi_d\) in Table \ref{tab1} (Biased tanh). 
We can see that in Fig. \ref{fig4}c, the spread of \(\bar{C}^\gamma_d\) is small for \(d=1\), and grows larger for \(d=2\) and \(d=3\). For \(d=4\) and \(d=5\), they have approximately the same sensitivity to \(\gamma\) as \(d=3\), which all happen to have very small \(\Phi_d\).

The analysis presented in this section can also be extended to multiplicative noise for simulating more realistic experimental scenarios.
However, unlike additive noise, the noise strength would depend on the state amplitude and is generally non-isotropic.
Therefore, the approximation of the noise floor energy (equation \ref{eq:noisefloor}) may no longer hold.

Another aspect to consider in physical RC is the nonlinearity.
In standard practice, it may be assumed that increasing the physical system's nonlinearity -- whether through a tunable parameter, or through adding more nonlinear elements -- increases the nonlinear computational capacity.
Although the reservoir may exhibit higher nonlinear capacity with additional nonlinearities, the corresponding representation energy may be too small to remain distinguishable from experimental noise, rendering the additional nonlinear capacity practically inaccessible.
As we have shown, for \(d>1\), a smaller nonlinear capacity which is well represented -- i.e. with a larger \(\Phi_{d}\) -- can be more favorable than a larger capacity with a small \(\Phi_{d}\).
\section{Conclusion}
In this work, we have shown how computation is organized inside a reservoir computer.
We applied this approach on the IPC to show the distribution of linear memory and nonlinear memory task representations in an ESN. 
As a result, we have demonstrated how the IPC tasks, grouped by degree, use up the reservoir resources from a state space geometry perspective. 
We have also elucidated the distinction between absolute computational capacities and their corresponding representation energy, shedding light on the resulting implications on physical RC.
The approach presented here can be used on any set of reconstruction tasks in general, and is not limited to IPC.

The results presented here suggest that the computational capabilities of RC are not governed solely by dimensionality expansion, but also by the degree-wise organization of task-relevant information across the reservoir state space.
A successful reservoir computer (for a given task) is that which can map task-relevant features to accessible dynamical modes (for example, in a given experimental setting).
A resourceful physical reservoir does this with the minimal amount of nodes -- whether real, or `virtual' in the time-delay RC sense -- without sacrificing task performance.

Finding smaller but equally performing reservoirs was the subject of an investigation using controllability matrices \cite{Whiteaker2022}.
Controllability matrices were also leveraged to show how linear reservoirs encode inputs in their states \cite{Verzelli2022}.
A comparison with the methods presented here could provide further insights on the representation ability of such reservoirs.
Future works may also leverage the analysis provided here to investigate the link between task representation and performance, apply this framework on experimental data, among other research directions.

\section*{Acknowledgements}
Conseil of Région Grand-Est and the European
Office of Aerospace Research and Development (EAORD) and AirForce Office for Scientific Research (AFOSR) via Grant FA8655-22-1-7031.
This work was supported partly by the french France 2030 program "Lorraine Initiative of Excellence", reference ANR-15- IDEX-04-LUE. 
This work was performed in part using computational resources from the “Ruche Mésocentre” computing center of Université Paris-Saclay, CentraleSupélec and École Normale Supérieure Paris-Saclay supported by CNRS and Région Île-de-France.

\section*{Conflict of Interest}
The authors declare no conflict of interest.

\section*{Author Contributions}
M.A.: Conceptualization, Methodology, Software, Validation, Formal analysis, Investigation, Visualization, Writing – original draft, Writing – review and editing.

D.R.: Validation, Writing – review and editing, Supervision, Funding acquisition.

\bibliographystyle{ieeetr}
\bibliography{bibliography}

\appendix

\section{Notes on the upper bound of the mode usage}
Each target can be written as a vector
\(\mathbf{z}_k \in \mathbb{R}^T\). The inner product between two target vectors is
\[
\langle \mathbf{z}_{k_1},\mathbf{z}_{k_2}\rangle_T
=
\frac{1}{T}\mathbf{z}_{k_1}^\top \mathbf{z}_{k_2}
=
\frac{1}{T}\sum_{t=1}^{T} z_{k_1}(t)z_{k_2}(t).
\]

In the limit $T \to \infty$, the empirical inner product converges to the theoretical expectation, which vanishes for distinct orthogonal targets:
\[
\langle \mathbf{z}_{k_1}, \mathbf{z}_{k_2} \rangle_T
\;\xrightarrow[T\to\infty]{}\;
\mathbb{E}\!\left[z_{k_1}(t) z_{k_2}(t)\right]
= 0,
\qquad k_1 \neq k_2 .
\]

Therefore, since the IPC targets form an orthogonal family in the limit $T\to\infty$, the bound on the mode usage follows directly from Bessel's inequality.

The scores \(s_{i,k}\) are squared projections of the per-mode normalized dynamics onto the targets.
For mode \(i\)
\[
S_i
=
\sum_k s_{i,k}
\leq 1
\quad
\text{as } T\to\infty .
\]
For finite sequences, the IPC targets are only approximately orthogonal, and therefore the empirical mode usage is expected to satisfy the approximate upper bound
\[
S_i \lesssim 1 .
\]

\end{document}